\documentclass{article}

\usepackage[nonatbib, final]{neurips_2021}
\usepackage{cite}           
\usepackage[utf8]{inputenc} 
\usepackage[T1]{fontenc}    
\usepackage{hyperref}       
\usepackage{url}            
\usepackage{booktabs}       
\usepackage{amsfonts}       
\usepackage{nicefrac}       
\usepackage{microtype}      
\usepackage{xcolor}         
\usepackage{graphicx}

\usepackage{amsmath}

\title{MAPIE: an open-source library for distribution-free uncertainty quantification}

\author{%
  Vianney Taquet$^1$, Vincent Blot$^1$, Thomas Morzadec$^1$, Louis Lacombe$^1$, Nicolas Brunel$^{1, 2}$ \\
  1: Quantmetry, 52, rue d'Anjou, 75008, Paris, France \\
  2: Laboratoire de Mathématiques et de Modélisation d'Evry, 
  ENSIIE, University Paris Saclay\\
}

\begin{document}

\maketitle

\begin{abstract}
Estimating uncertainties associated with the predictions of Machine Learning (ML) models is of crucial importance to assess their robustness and predictive power. 
In this submission, we introduce MAPIE (Model Agnostic Prediction Interval Estimator), an open-source Python library that quantifies the uncertainties of ML models for single-output regression and multi-class classification tasks.
MAPIE implements conformal prediction methods, allowing the user to easily compute uncertainties with strong theoretical guarantees on the marginal coverages and with mild assumptions on the model or on the underlying data distribution.
MAPIE is hosted on scikit-learn-contrib and is fully "scikit-learn-compatible". As such, it accepts any type of regressor or classifier coming with a scikit-learn API. The library is available at: \url{https://github.com/scikit-learn-contrib/MAPIE/}.
\end{abstract}

\section{Introduction}
\label{introduction}

Quantifying the uncertainties of ML model predictions is of crucial importance for developing and deploying reliable artificial intelligence (AI) systems. Uncertainty quantification (UQ) involves all the stakeholders who develop and use AI models.
First, UQ allows the designers of AI systems to better understand the predictive power of their model and assess the validity of the model predictions on new data points.
Second, UQ allows AI operators, such as business stakeholders, to optimize the risk management when making business decisions based on AI system predictions.
Third, UQ helps AI regulators to assess the compliance of the AI system with the regulation in force.
Fourth, UQ allows the AI systems to be more transparent and trustworthy for people impacted by the decisions made from AI.

There is therefore a strong need for libraries of uncertainty quantification that respect three fundamental pillars.
First, implemented methods have to be model and use case agnostic in order to address all relevant use cases tackled in industry, such as natural language processing, time series, or computer vision, using state-of-the-art ML models, like neural networks or gradient boosting models. 
Second, methods must have strong theoretical guarantees at least on the marginal coverage (and possibly on the conditional coverage) of the estimated uncertainties with as little assumption on the data or the model as possible. This ensures AI system designers, operators, and regulators to be confident about the predictions provided by the models.
Third, libraries need to be open-source 
and respect state-of-the-art programming standards to develop trustworthy AI systems.

Resampling or undersampling methods have been used for a few decades to estimate the robustness of predictions \cite{Quenouille1956, Efron1979}.
Among other, Bootstrap is probably the most commonly used methods since it is easy to implement and allow the user to easily obtain confidence intervals associated with the predictions. 
However, the standard jackknife technique suffers from instabilities in some cases and can be unusable on practical use cases when the number of training data samples is high. 

Although modern resampling techniques have been implemented in recent R packages (see for instance \cite{ryantibs}), they are not yet implemented in Python and in particular within the standard scikit-learn framework \cite{scikit-learn}. 
Some scikit-learn regressors allow the users to estimate confidence intervals associated with the model predictions but they are either simple regressors, such as the Bayesian Ridge model, or based on quantile regression for gradient boosting.
IBM's UQ360 \cite{UQ360} is noteworthy as this library aims at incorporating several complementary UQ methods (Bayesian inference, quantile regression, jackknife, etc). However, it is still in its early stages and has been inactive for the past 6 months.

Since 2021, we develop the MAPIE (Model Agnostic Prediction Interval Estimator) library in order to address the three aforementioned pillars. MAPIE is an open-source Python library hosted on scikit-learn-contrib. It follows the scikit-learn guidelines; the only technical requirement is to have a scikit-learn API and accepts base scikit-learn-compatible estimators. Importantly, MAPIE implements conformal prediction methods for regression and classification settings and is therefore model and (soon to be) use case agnostic. Conformal prediction methods allow MAPIE to have mathematical guarantees on the marginal coverages on the uncertainties.

Section \ref{presentation} presents the conformal prediction methods implemented in MAPIE. Section \ref{practice} describes MAPIE in practice by listing the main input parameters and how to use them. Section \ref{examples} presents two illustrative examples. Section \ref{perspectives} concludes with our perspectives.

\section{Methods implemented in MAPIE}
\label{presentation}

\subsection{General settings}

Before describing the methods, we briefly present the mathematical setting. For a regression or a classification problem in a standard independent and identically distributed (i.i.d) case, our training data $(X, Y) = \{(x_1, y_1), \ldots, (x_n, y_n)\}$ has an unknown distribution $P_{X, Y}$. 
For any risk level $\alpha$ between 0 and 1, we aim at constructing a prediction interval or a prediction set $\hat{C}_{n, \alpha}(X_{n+1})$ for a new
observation $\left( X_{n+1},Y_{n+1} \right)$ 
such that : 

\begin{equation}
    P \{Y_{n+1} \in \hat{C}_{n, \alpha}(X_{n+1}) \} \geq 1 - \alpha
\end{equation}

In words, for a typical risk level $\alpha$ of $10 \%$, we want to construct prediction intervals or prediction sets that contain the true observations for at least $90 \%$ of the new test data points. 
To achieve this objective, MAPIE includes several split- or cross-conformal methods for constructing prediction intervals or prediction sets for regression or classification tasks, respectively. We can sum up UQ with split-conformal prediction as follows : 
\begin{itemize}
\item Choose a "conformity score" that quantifies how well the observation conforms with the prediction of the ML model. The higher the value, the more atypical the point. For regression, the conformity scores can be the residuals. For classification, they can be the cumulative sum of ranked softmax scores.
\item  Train the model on a training set and calibrate the conformity scores on a calibration set apart from the training set to avoid over-fitting.
\item  Estimate the quantile of the conformity score distribution associated with the chosen risk level $\alpha$.
\item  Construct prediction intervals or prediction sets for new test points based on this quantile.
\end{itemize}

\subsection{MAPIE for tabular regression}

MAPIE has three state-of-the-art conformal prediction methods (along with their derivatives) for tabular regression : Jackknife+, CV+ and Jackknife+-after-Bootstrap.

\paragraph{Jackknife+.}
Presented in \cite{Barber2021}, the Jackknife+ method is based on the standard Jackknife, which constructs a set of {\it leave-one-out} models. 
Estimating the prediction intervals is carried out in three main steps.
First, train $n$ leave-one-out models (one for each training instance), each leave-one-out model is trained on the entire training set except the corresponding training point.
Second, compute the corresponding conformity scores (here, the leave-one-out residuals)
$|Y_i - \hat{\mu}_{-i}(X_i)|$. 
Third, fit the regression function $\hat{\mu}$ on the entire training set and construct the prediction intervals from the distribution of the computed leave-one-out residuals and the desired $1 - \alpha$ quantile.
This method avoids over-fitting but still does not guarantee the targeted coverage if $\hat{\mu}$ is unstable, for example when the sample size is close to the number of features \cite{Barber2021}.

Unlike the standard jackknife method which returns a prediction interval centered around the prediction of the model trained on the entire dataset, the so-called Jackknife+ method also uses the predictions from all leave-one-out models on the new test point to take the variability of the regression function into account.
The resulting confidence interval can therefore be summarized as follows

\begin{equation}
\hat{C}_{n, \alpha}^{\rm jackknife+}(X_{n+1}) = [ \hat{q}_{n, \alpha}^-\{\hat{\mu}_{-i}(X_{n+1}) - R_i^{\rm LOO} \}, \hat{q}_{n, \alpha}^+\{\hat{\mu}_{-i}(X_{n+1}) + R_i^{\rm LOO} \}] 
\end{equation}

As described in \cite{Barber2021}, this method guarantees a higher stability with a coverage level of $1-2\alpha$ for a target coverage level of $1-\alpha$, under the assumption of independence of the distribution of the data $(X, Y)$.

\paragraph{CV+.}
In order to reduce the computational cost, one can adopt a cross-validation approach instead of a leave-one-out approach, with the so-called CV+ method.
Similar to the Jackknife+ method, estimating the prediction intervals with CV+ is performed in four main steps. First, split the training set into $K$ disjoint subsets $S_1, S_2, ..., S_K$ of equal size. Second, regression functions $\hat{\mu}_{-S_k}$ are fitted on the training set with the corresponding $k^{th}$ fold removed. Third, the corresponding out-of-fold residuals are computed for each $i^{th}$ point 
  $|Y_i - \hat{\mu}_{-S_{k(i)}}(X_i)|$ where $k(i)$ is the fold containing $i$.
Finally, similar to Jackknife+, the regression functions $\hat{\mu}_{-S_{k(i)}}(X_i)$  are used to estimate the prediction intervals. 

As for Jackknife+, this method guarantees a coverage level higher than $1-2\alpha$ for a target coverage level of $1-\alpha$, assuming essentially the independence of the data.
As pointed out by \cite{Barber2021}, Jackknife+ can be considered as a special case of CV+ with $K = n$. 
In practice, this method results in slightly wider prediction intervals and is therefore likely more conservative, but gives a reasonable compromise for large datasets when the Jacknife+ method is unfeasible.

\paragraph{Jackknife+-after-Bootstrap.}
An alternative way to reduce the computational cost 
is to adopt a bootstrap approach instead of cross-validation, called the Jackknife+-after-bootstrap method, offered by \cite{Kim2020}.
Similar to CV+, estimating the prediction intervals with Jackknife+-after-bootstrap is performed in four main steps. First, resample the training set with replacement (bootstrap) $K$ times, to get the (non disjoint) bootstraps $B_{1},..., B_{K}$ of equal size. Second, fit $K$ regressions functions $\hat{\mu}_{B_{k}}$ on the bootstraps ($B_{k}$), and compute the predictions on the complementary sets $B_k^c$. Third, aggregate these predictions according to a given aggregation function, typically {\rm mean} or {\rm median}, and compute the residuals $|Y_j - {\rm agg}(\hat{\mu}(B_{K(j)}(X_j)))|$ are computed for each $X_j$ (with $K(j)$ the boostraps not containing $X_j$).
The sets $\{{\rm agg}(\hat{\mu}_{K(j)}(X_i) + r_j\}$ (where $j$ indexes the training set) are used to estimate the prediction intervals.

As for Jackknife+, this distribution-free method guarantees a coverage level higher than $1 - 2\alpha$ for a target coverage level of $1 - \alpha$.

\subsection{MAPIE for Time Series}

The methods implemented in MAPIE for tabular regression assume that data are exchangeable. This hypothesis is hardly reasonable for dynamical time series, thus requiring a specific method. The method implemented in MAPIE is based on the algorithm {\it Ensemble Batch Prediction Intervals} (EnbPI, see \cite{Xu2021}). In this method, the assumption on data exchangeability is replaced with assumptions on the residuals of the estimators (errors process and estimations quality). Moreover the coverage guaranty is not absolutely but approximately valid up to these assumptions validity.

The corresponding algorithm is very closed to Jackknife+-after-Bootstrap. It is notably based on bootstrapping (with a block bootstrap that suits to time series). The predictions are the aggregations of the predictions of the refitted estimators. The bounds of the prediction intervals are the predictions plus or minus suitable quantiles of the conformity scores. So the widths of the intervals only depend on the conformity scores and not on the variability of the predictions of the refitted estimators, contrary to Jackknife+. 

The conformity scores are no longer the absolute values of the residuals but the relative values. So the prediction intervals are not symmetric with respect to the predictions. Moreover, the levels of the quantiles are optimized so that the widths of the intervals are minimum while the difference between the quantiles' level is $1 - \alpha$. 

Finally, the conformity scores are updated during the prediction process. So they can dynamically take into account, for example, an increase in the variance of the residuals or a model deterioration.

\subsection{MAPIE for classification}

Three methods for multi-class classification UQ have been implemented in MAPIE so far : LABEL \cite{Sadinle2019}, Adaptive Prediction Sets \cite{Romano2020} and Top-K \cite{Angelopoulos2020}. 
The difference between these methods is the way the conformity scores are computed.

\paragraph{LABEL.}
In the LABEL method, the conformity score is defined as as one minus the score of the true label. For each point $i$ of the calibration set : 

\begin{equation}
s_i(X_i, Y_i) = 1 - \hat{\mu}(X_i)_{Y_i}
\end{equation}

Once the conformity scores ${s_1, ..., s_n}$ are estimated for all calibration points, we compute the $(n+1)*(1-\alpha)/n$ quantile $\hat{q}$ as follows : 

\begin{equation} \label{eq:score}
    \hat{q} = Quantile \left(s_1, ..., s_n ; \frac{\lceil(n+1)(1-\alpha)\rceil}{n}\right) \\
\end{equation}

Finally, we construct a prediction set by including all labels with a score higher than the estimated quantile :

\begin{equation*}
    \hat{C}(X_{test}) = \{y : \hat{\mu}(X_{test})_y \geq 1 - \hat{q}\}
\end{equation*}

This simple approach allows us to construct prediction sets coming with a theoretical guarantee on the marginal coverage. However, although this method generally results in small prediction sets, it tends to produce empty ones when the model is uncertain, for example at the border between two classes.





\paragraph{APS.}
The so-called Adaptive Prediction Set (APS) method overcomes the problem encountered by the LABEL method through the construction of prediction sets which are by definition non-empty. The conformity scores are computed by summing the ranked scores of each label, from the higher to the lower until reaching the true label of the observation :

\begin{equation*}
   s_i(X_i, Y_i) = \sum^k_{j=1} \hat{\mu}(X_i)_{\pi_j} \quad \text{where} \quad Y_i = \pi_j 
\end{equation*}

The quantile $\hat{q}$ is then computed the same way as the score method. For the construction of the prediction sets for a new test point, the same procedure of ranked summing is applied until reaching the quantile, as described in the following equation : 


\begin{equation*}
   \hat{C}(X_{test}) = \{\pi_1, ..., \pi_k\} \quad \text{where} \quad k = \text{inf}\{k : \sum^k_{j=1} \hat{\mu}(X_{test})_{\pi_j} \geq \hat{q}\}
\end{equation*}

By default, the label whose cumulative score is above the quantile is included in the prediction set. However, its incorporation can also be chosen randomly based on the difference between its cumulative score and the quantile so the effective coverage remains close to the target (marginal) coverage. We refer the reader to \cite{Romano2020} and \cite{Angelopoulos2020} for more details about this aspect.

\paragraph{Top-K.}
Introduced by \cite{Angelopoulos2020}, the specificity of the Top-K method is that it will give the same prediction set size for all observations. The conformity score is the rank of the true label, with scores ranked from higher to lower. The prediction sets are build by taking the $\hat{q}^{th}$ higher scores. The procedure is described in equation \ref{eq:topk}. 

\begin{equation*}
   s_i(X_i, Y_i) = j \quad \text{where} \quad Y_i = \pi_j \quad \text{and} \quad \hat{\mu}(X_i)_{\pi_1} > ... > \hat{\mu}(X_i)_{\pi_j} > ... > \hat{\mu}(X_i)_{\pi_n}
\end{equation*}

\begin{equation} \label{eq:topk}
    \hat{q} = \left \lceil Quantile \left(s_1, ..., s_n ; \frac{\lceil(n+1)(1-\alpha)\rceil}{n}\right) \right\rceil
\end{equation}

\begin{equation*}
   \hat{C}(X_{test}) = \{\pi_1, ..., \pi_{\hat{q}}\} 
\end{equation*}

Finally, it should be noted that MAPIE includes split- and cross-conformal strategies for the LABEL and APS methods, but only the split-conformal one for Top-K.

\section{MAPIE in practice}
\label{practice}

\begin{figure}[h!]
  \centering
  \includegraphics[width=12cm]{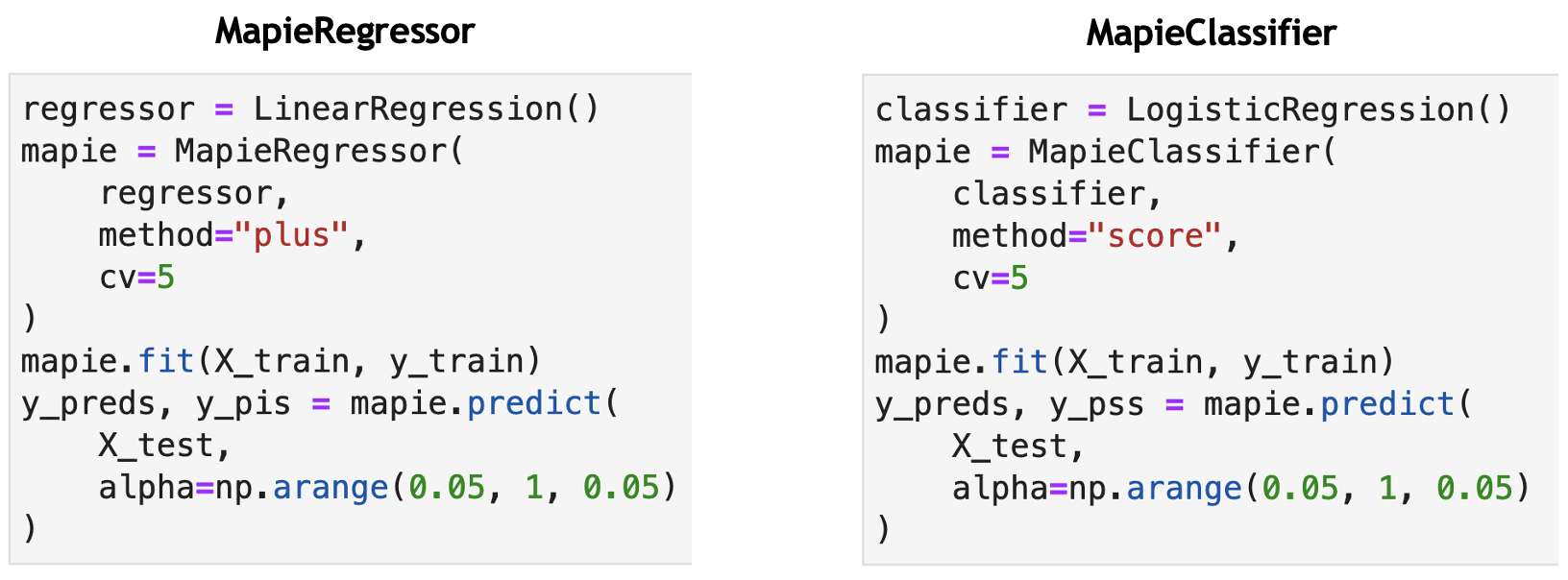}
  \caption{Description of the Python commands needed for estimating prediction intervals and prediction sets with \texttt{MapieRegressor} and \texttt{MapieClassifier}, respectively.}
  \label{mapie_workflow}
\end{figure}

The MAPIE library offers two base Pythonic classes, called \texttt{MapieRegressor} and \texttt{MapieClassifier} for estimating prediction intervals and prediction sets, respectively, via the construction of conformity scores. Figure \ref{mapie_workflow} presents the typical commands needed for quantifying the uncertainties. As the MAPIE base classes are inherited from scikit-learn \texttt{BaseEstimator} classes, the API is very intuitive to anyone familiar with scikit-learn and follows a initialization-fit-predict process. 

After initializing the base scikit-learn-compatible base model, one needs to initialize the desired MAPIE class with two main arguments that define the strategy for uncertainty quantification :

\begin{itemize}

 \item "cv" defines the train / calibration set splitting strategy for training the model and calibrating the conformity scores. It can be "prefit" in the split-conformal case where the base model is already fitted on a given training set while the set given to MAPIE is used directly for calibrating the conformity scores. For cross-conformal methods, one can simply define an integer that sets the number of splits and MAPIE will call internally the corresponding \texttt{BaseCrossValidator} object, such as \texttt{LeaveOneOut} and \texttt{KFold} for the Jackknife or CV strategies, respectively.
 
 \item "method" controls the strategy for constructing the prediction intervals or prediction sets. For regression tasks, one can choose among "base", "plus", or "minmax". For example, \texttt{method="plus"} together with \texttt{cv=KFold(5)} defines the CV+ method with 5 folds. For classification, one can choose "score" for the LABEL method from \cite{Sadinle2019}, "cumulated\_score" for the Adaptive Prediction Set (APS) method by \cite{Romano2020}, and "top\_k" for the Top-K method  by \cite{Angelopoulos2020}. For example, \texttt{method="cumulated\_score"} together with \texttt{cv=KFold(5)} defines the APS method with the CV+ strategy as defined in Algorithm 2 of \cite{Romano2020}.
\end{itemize}

Methods that deviate from these standard cases (i.e. training and test i.i.d. datasets sampled from similar distribution) need to be implemented in other classes using \texttt{MapieRegressor} or \texttt{MapieClassifier} as base classes. 
For instance, the EnbPI method is implemented in the \texttt{MapieTimeSeriesRegressor} class that inherits from \texttt{MapieRegressor}.
This process allows anyone to suggest an implementation of a new conformal prediction method through a dedicated Pull Request that follows the MAPIE guidelines without modifying the base classes.

\section{Examples}
\label{examples}

We present two simple examples of uncertainty quantification on time series and computer vision settings using MAPIE. The notebooks for both examples can be found in the github repository \url{https://github.com/scikit-learn-contrib/MAPIE/}. 

\subsection{MAPIE for Time series}

We illustrate the EnbPI algorithm with \texttt{MapieTimeSeriesRegressor} on the Victoria electricity demand dataset with an artificial change point on the test set.
The Victoria electricity demand dataset consists in the hourly demand of electricity in the Victoria state in Australia between January 1st and February 23rd of 2014. The test set is the last week, and the remaining weeks are included in the training set. We added a sudden decrease of the electricity demand of 2 GW on February 22 to simulate a change point in the test set. 
The explanatory variables are the lags of the demand up to 5 previous hours and the temperature. 
The base ML model is a Random Forest whose hyperparameters are optimized through chronological cross-validation. We then use this tuned Random Forest as the base model for the EnbPI method, using 100 block bootstrap resamplings and with a block length of 48 hours.
We sequentially estimate the prediction interval on the electricity demand one step ahead. 

\begin{figure}[h!]
  \centering
  \includegraphics[width=12cm]{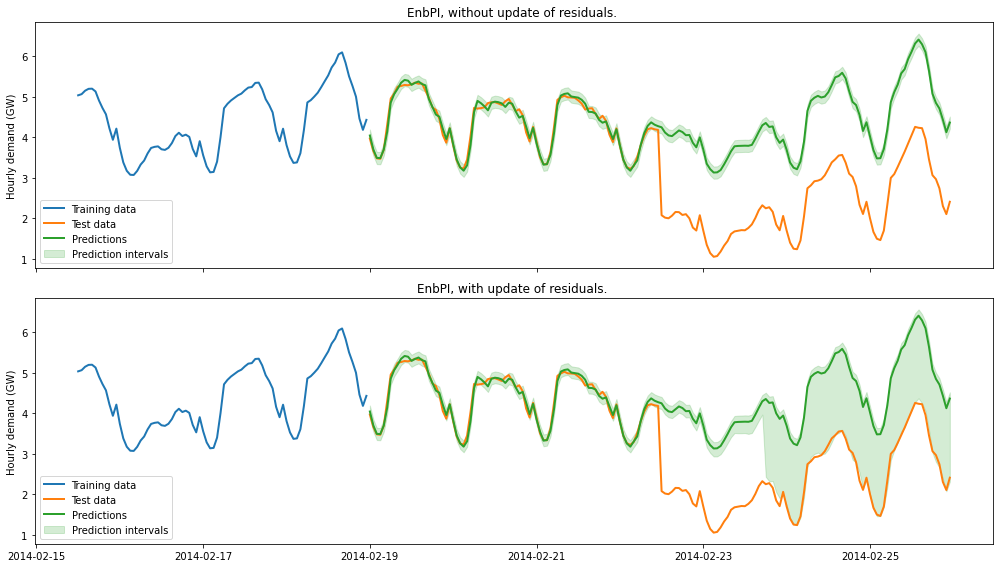}
  \caption{Predictions obtained by \texttt{MapieTimeSeriesRegressor} without (top) and with (bottom) partial update of residuals for computing prediction intervals from a Random Forest regressor as based model trained on the Australian electricity dataset.}
  \label{enbpi}
\end{figure}

Figure \ref{enbpi} compares the estimated prediction intervals without and with update of the residuals, that is a key point of the EnbPI method. The training data do not contain a change point, hence the base model cannot anticipate it.
Without update of the residuals, the prediction intervals are built upon the distribution of the residuals of the training set. Therefore they do not cover the true observations after the change point, leading to a sudden decrease of the coverage.
However, the partial update of the residuals allows the method to capture the increase of uncertainties of the model predictions. One can notice that the uncertainty's explosion happens about one day late. This is because enough new residuals are needed to change the quantiles obtained from the residuals distribution.

\subsection{MAPIE for image classification} 

We illustrate uncertainty quantification on image classification with the famous CIFAR10 dataset which consists in images belonging to 10 classes : airplane, automobile, bird, cat, deer, dog, frog, horse, ship and truck.

As mentioned before, MAPIE is a "scikit-learn compatible" library, meaning that the base classifier must include the \texttt{classes\_}, \texttt{trained\_} attributes  and the \texttt{fit}, \texttt{train}, \texttt{predict}, \texttt{predict\_proba}, \texttt{\_\_sklearn\_is\_fitted\_\_} methods in order to be accepted by \texttt{MapieClassifier}.
Hence, for computer vision settings, one can create a wrapper around a deep learning model object (or any other non-scikit-learn compatible ML model) to be digested by \texttt{MapieClassifier}. An example of such a wrapper can be found in the Cifar10 notebook of the MAPIE repository \footnote{\url{https://github.com/scikit-learn-contrib/MAPIE/blob/master/notebooks/classification/Cifar10.ipynb}}.

\begin{figure}[h!]
  \centering
  \includegraphics[width=\textwidth]{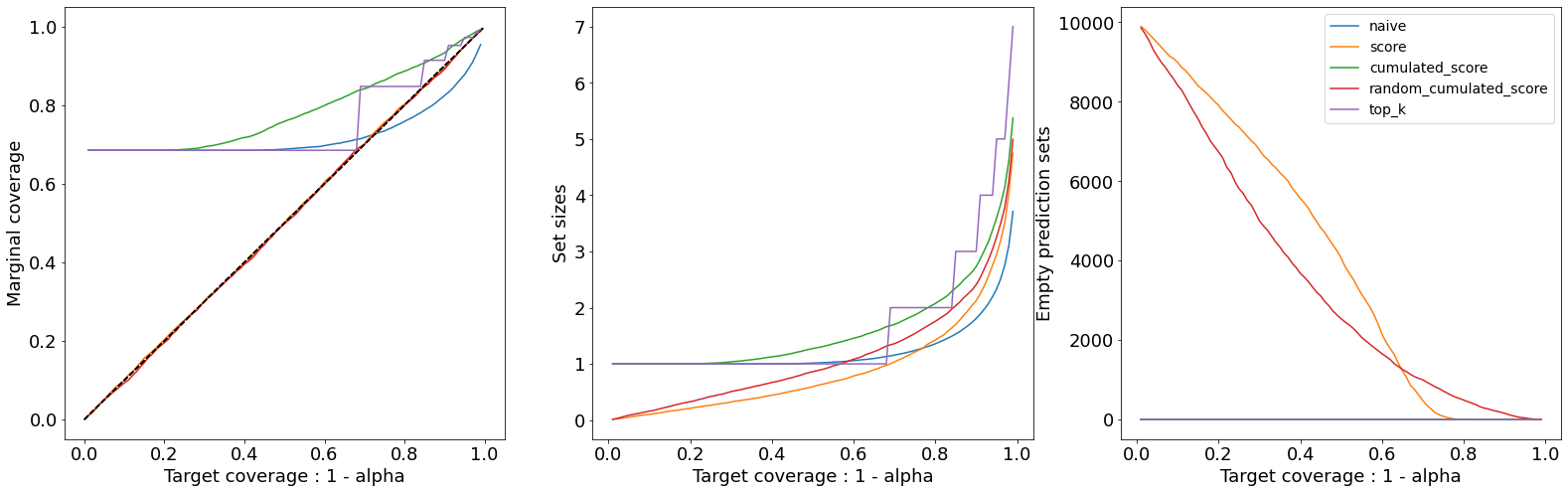}
  \caption{Comparison of the number of empty prediction sets, marginal coverages and average prediction set sizes for the different classification methods implemented in \texttt{MapieClassifier}.}
  \label{cifar}
\end{figure}

After training a small convolutional neural network on the CIFAR10 dataset, we used this wrapper to fit MAPIE on a calibration set and create prediction sets on a test set. Figure \ref{cifar} compares the different aforedescribed methods implemented in MAPIE with a naive one which directly includes all the labels such that their sum is just above the target coverage level. 

As expected, the "naive" method has a coverage which is below the target coverage for $\alpha$ values higher than 0.6 : this is because output scores of the model do not represent probabilities. Thus, even by adding the labels, there are no guarantees that the coverage will be achieved. Other methods, on the other hand, all achieve the required coverage regardless the $\alpha$ value because they calibrate the scores with a dataset not seen by the model during training.

Even though the methods all achieve the required coverages, the averaged sizes of their prediction sets are quite different, especially at low $\alpha$ values. As discussed in the previous section, the "score" method achieves the lowest averaged size for all target coverages compared to the "cumulated\_score" and "top\_k" methods.
However, the "score" method may also result in empty prediction sets. This situation arises in uncertain cases where the predicted scores of all labels do not reach the target quantile.
On the second hand, the "cumulated\_score", by definition, always includes at least one label, the one whose cumulated score is higher than the quantile, but induces over-estimated marginal coverages for low $\alpha$ values. To force the marginal coverage to stay close to the target one, one can choose the "random\_cumulated\_score" which includes randomly the last label. The major drawback of the latter method is that it creates empty prediction sets which indicate low uncertainties, unlike the "score" method.

\section{Future works}
\label{perspectives}

\subsection{Current implementations}

In addition to the extension of MAPIE to time series settings, other methods derived from standard i.i.d. split-/cross-conformal paradigms are being implemented, such as conformalized quantile regression \cite{Romano2019} or covariate shift \cite{Tibshirani2019}. 

\paragraph{Conformalized Quantile Regression (LLA).} 
Where all the previously mentioned methods from \texttt{MapieRegressor} guarantee the targeted coverage, the width does not vary locally and therefore fails to undertake heteroscedastic noise. 
Through the implementation of Conformalized Quantile Regression \cite{Romano2019}, we created a new 
class \texttt{MapieQuantileRegressor} that offers a new "quantile" method which adapts the width of the prediction intervals to the noise in the data. 

\subsection{A framework for straightforward implementation of new conformal prediction methods}

Motivated by the implementation of various conformal prediction methods, we are also working on the refactorization of the base \texttt{MapieRegressor} and \texttt{MapieClassifier} class in order to offer a more robust framework for optimizing the implementation of new conformal prediction algorithms by external collaborators. 

Future implementations that would benefit from this refactorization could adaptive conformal inference \cite{Gibbs2020}, multi-target regression through copula-based conformal prediction \cite{Messoudi2021} for regression settings or regularized adaptive prediction sets \cite{Angelopoulos2020} or conformal label shift \cite{Podkopaev2021} for classification settings.

\subsection{Extension to new settings}

Finally, we also plan to extend MAPIE to other computer vision settings, such as image segmentation or object detection for instance through the implementation of the Risk-controlling Prediction Sets \cite{Bates2021} or the Learn Then Test \cite{Angelopoulos2021} frameworks.

\section*{Acknowledgement}

We would like to thank Quantmetry and Région Ile-de-France for their financial support, as well as the Michelin AI R\&D team for enlightening discussions.

\bibliographystyle{unsrt}
\bibliography{biblio.bib}

\begin{thebibliography}{10}

\bibitem{Quenouille1956}
M.~H. Quenouille.
\newblock Notes on bias in estimation.
\newblock {\em Biometrika}, 43(3/4):353--360, 1956.

\bibitem{Efron1979}
B.~Efron.
\newblock Bootstrap methods: Another look at the jackknife.
\newblock {\em The Annals of Statistics}, 7(1):1--26, 1979.

\bibitem{ryantibs}
\url{https://github.com/ryantibs/conformal}.
\newblock Accessed: 2021-06-10.

\bibitem{scikit-learn}
F.~Pedregosa, G.~Varoquaux, A.~Gramfort, V.~Michel, B.~Thirion, O.~Grisel,
  M.~Blondel, P.~Prettenhofer, R.~Weiss, V.~Dubourg, J.~Vanderplas, A.~Passos,
  D.~Cournapeau, M.~Brucher, M.~Perrot, and E.~Duchesnay.
\newblock Scikit-learn: Machine learning in {P}ython.
\newblock {\em Journal of Machine Learning Research}, 12:2825--2830, 2011.

\bibitem{UQ360}
\url{https://github.com/IBM/UQ360}.
\newblock Accessed: 2022-05-30.

\bibitem{Barber2021}
R.~Foygel~Barber, E.~J. Candès, A.~Ramdas, and R.~J. Tibshirani.
\newblock Predictive inference with the jackknife+.
\newblock {\em Ann. Statist.}, 49(1):486--507, 02 2021.

\bibitem{Kim2020}
B.~Kim, C.~Xu, and Rina Barber.
\newblock Predictive inference is free with the jackknife+-after-bootstrap.
\newblock In H.~Larochelle, M.~Ranzato, R.~Hadsell, M.F. Balcan, and H.~Lin,
  editors, {\em Advances in Neural Information Processing Systems}, volume~33,
  pages 4138--4149. Curran Associates, Inc., 2020.

\bibitem{Xu2021}
C.~Xu and Y.~Xie.
\newblock Conformal prediction for dynamic time-series, 2020.

\bibitem{Sadinle2019}
Sadinlem M., J.~Lei, and L.~Wasserman.
\newblock Least ambiguous set-valued classifiers with bounded error levels.
\newblock {\em Journal of the American Statistical Association},
  114(525):223--234, 2019.

\bibitem{Romano2020}
Y.~Romano, M.~Sesia, and E.~Candes.
\newblock Classification with valid and adaptive coverage.
\newblock In H.~Larochelle, M.~Ranzato, R.~Hadsell, M.F. Balcan, and H.~Lin,
  editors, {\em Advances in Neural Information Processing Systems}, volume~33,
  pages 3581--3591. Curran Associates, Inc., 2020.

\bibitem{Angelopoulos2020}
A.~Angelopoulos, S.~Bates, J.~Malik, and M.~I. Jordan.
\newblock Uncertainty sets for image classifiers using conformal prediction.
\newblock 2020.

\bibitem{Romano2019}
Y.~Romano, E.~Patterson, and E.~Candes.
\newblock Conformalized quantile regression.
\newblock In H.~Wallach, H.~Larochelle, A.~Beygelzimer, F.~d\textquotesingle
  Alch\'{e}-Buc, E.~Fox, and R.~Garnett, editors, {\em Advances in Neural
  Information Processing Systems}, volume~32. Curran Associates, Inc., 2019.

\bibitem{Tibshirani2019}
R.~J. Tibshirani, R.~Foygel Barber, E.~J. Candes, and A.~Ramdas.
\newblock Conformal prediction under covariate shift, 2019.

\bibitem{Gibbs2020}
I.~Gibbs and E.~Candes.
\newblock Adaptive conformal inference under distribution shift.
\newblock In M.~Ranzato, A.~Beygelzimer, Y.~Dauphin, P.S. Liang, and J.~Wortman
  Vaughan, editors, {\em Advances in Neural Information Processing Systems},
  volume~34, pages 1660--1672. Curran Associates, Inc., 2021.

\bibitem{Messoudi2021}
S.~Messoudi, S.~Destercke, and S.~Rousseau.
\newblock Copula-based conformal prediction for multi-target regression, 2021.

\bibitem{Podkopaev2021}
A.~Podkopaev and A.~Ramdas.
\newblock Distribution-free uncertainty quantification for classification under
  label shift, 2021.

\bibitem{Bates2021}
S.~Bates, A.~Angelopoulos, L.~Lei, J.~Malik, and M.~Jordan.
\newblock Distribution-free, risk-controlling prediction sets.
\newblock {\em Journal of the ACM (JACM)}, 68(6):1--34, 2021.

\bibitem{Angelopoulos2021}
A.~N Angelopoulos, S.~Bates, E.~J Cand{\`e}s, M.~I. Jordan, and L.~Lei.
\newblock Learn then test: Calibrating predictive algorithms to achieve risk
  control.
\newblock {\em arXiv preprint arXiv:2110.01052}, 2021.

\end{thebibliography}

\end{document}